\documentclass[letterpaper, 10 pt, conference]{ieeeconf}  

\IEEEoverridecommandlockouts                              

\usepackage{xcolor}
\usepackage{graphicx}
\usepackage{pifont}

\global\long\def\mymatrix#1{\boldsymbol{#1}}%

\global\long\def\myvec#1{\boldsymbol{#1}}%

\newcommand{\joints}{\myvec q}

\IEEEoverridecommandlockouts                              

\usepackage[utf8]{inputenc}
\usepackage{leftidx}
\usepackage{bm}
\usepackage{amssymb,amsmath,amsfonts,empheq}
\usepackage{graphicx}
\usepackage{siunitx}
\usepackage{multicol}
\usepackage{algorithm}
\usepackage[noend]{algpseudocode}
\usepackage{type1cm}
\usepackage{pifont}
\usepackage{lettrine}
\usepackage{color}
\usepackage{cite}
\usepackage{breqn}
\usepackage{tikz}
\usepackage{array}
\usepackage{import}
\usepackage{amsmath}
\usepackage{amssymb}
\usepackage{bm} 
\usepackage{booktabs}
\usepackage{optidef}
\usepackage{graphicx}
\usepackage{xcolor}
\usepackage{siunitx}
\usepackage{fixltx2e}
\usepackage{multirow}
\usepackage{cite}
\usepackage{import}
\usepackage[draft]{hyperref}
\usepackage{microtype}
\usepackage{tabularx}
\usepackage{booktabs}
\usepackage{comment}
 \usepackage{microtype}
 \usepackage{xifthen}

\usepackage{multirow}
\usepackage{scalerel}
\usepackage{fixltx2e}
\usepackage{amsmath}
\usepackage{dsfont}

\makeatletter

\renewcommand{\baselinestretch}{0.96}
\title{\LARGE \bf
Coupling Tensor Trains with Graph of Convex Sets: Effective Compression, Exploration, and Planning in the C-Space
}

\author{Gerhard Reinerth$^{1}$, Riddhiman Laha$^{1,2}$, and Marcello Romano$^{1}$
\thanks{}
\thanks{$^{1}$Gerhard Reinerth, Riddhiman Laha, and Marcello Romano are with the
        Technical University of Munich, Germany
        {\tt\small g.reinerth@tum.de, marcello.romano@tum.de}}%
\thanks{$^{2}$Riddhiman Laha is also with Northeastern University,
        Boston, USA
        {\tt\small r.laha@northeastern.edu}}%
}

\renewcommand{\baselinestretch}{0.94}
\begin{document}

\maketitle
\thispagestyle{empty}
\pagestyle{empty}

\begin{abstract}
We present TANGO (Tensor ANd Graph Optimization), a novel motion planning framework that integrates tensor-based compression with structured graph optimization to enable efficient and scalable trajectory generation. While optimization-based planners such as the Graph of Convex Sets (GCS) offer powerful tools for generating smooth, optimal trajectories, they typically rely on a predefined convex characterization of the high-dimensional configuration space—a requirement that is often intractable for general robotic tasks. TANGO builds further by using Tensor Train decomposition to approximate the feasible configuration space in a compressed form, enabling rapid discovery and estimation of task-relevant regions. These regions are then embedded into a GCS-like structure, allowing for geometry-aware motion planning that respects both system constraints and environmental complexity. By coupling tensor-based compression with structured graph reasoning, TANGO enables efficient, geometry-aware motion planning and lays the groundwork for more expressive and scalable representations of configuration space in future robotic systems. Rigorous simulation studies on planar and real robots reinforce our claims of effective compression and higher quality trajectories.

\end{abstract}

\section{Introduction}
The problem of generalized motion planning for articulated systems, particularly under combined system and task-level costs and constraints, remains an open and active area of research in robotics~\cite{kingston2018sampling}. In order to make it tractable, the following sub-areas have been studied: (a) Task assignment~\cite{richards2005mixed, alighanbari2003coordination,zhang2017modelling}, (b) Path and trajectory planning~\cite{cetin2007hybrid, deits2014footstep}, and (c) Collision avoidance~\cite{deits2015computing, laha2021reactive, laha2023predictive, werner2024approximating, werner2024faster}. Notwithstanding, region generation in the C-space remains a roadblock~\cite{von2024using}. Researchers have proposed various perspectives to tackle this challenge. Decomposition methods for handling non-convex shapes vary significantly in strategy. Iterative approaches, such as that of Lien and Amato~\cite{lien2004approximate}, repeatedly partition the shape by removing the largest concavity, gradually improving convexity. Optimization-based methods, like Liu et al.~\cite{liu2010convex}, instead formulate the problem as a mixed-integer program, identifying cutting planes that minimize concavity under a prescribed threshold. Clustering-based techniques, exemplified by~\cite{mamou2009simple}, group faces of the shape into clusters that together approximate convex components. While differing in methodology—iterative refinement, optimization-driven cuts, or face clustering—all of these techniques return only approximately convex partitions that cover the original geometry. A key drawback is that using the convex hulls of these components in motion planning may inflate feasible regions and inadvertently intersect with obstacles, undermining safety guarantees. 

\begin{figure}[t]
    \centering
    \includegraphics[width=1\linewidth]{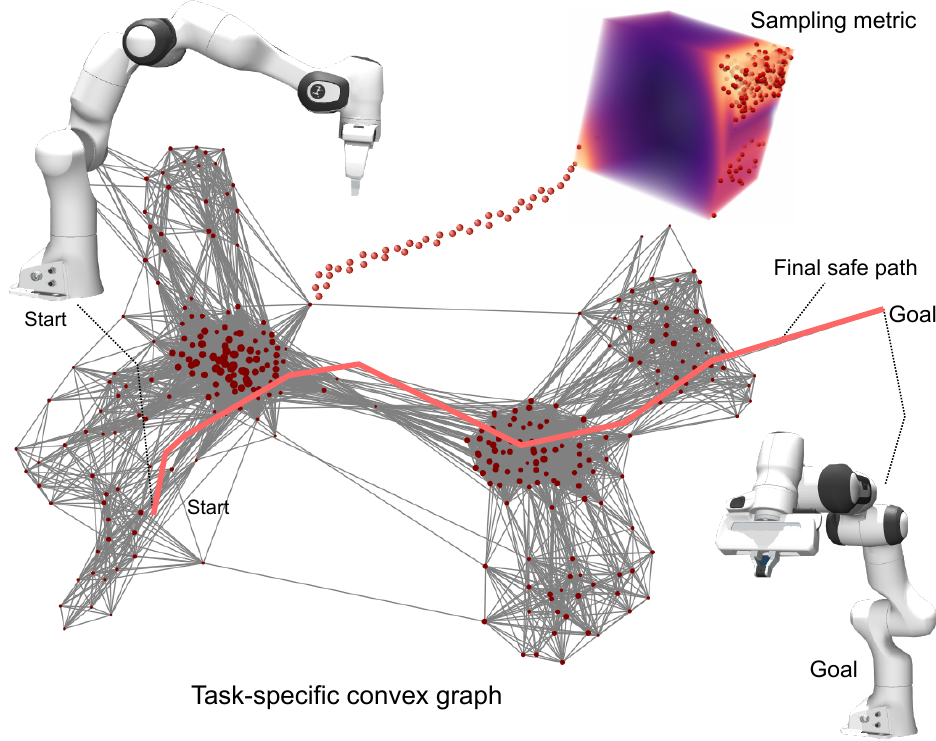}
    \caption{Our algorithm enables a task-specific sampling metric for approximating the feasible configuration space for a system. These approximated regions are then used for discovering a convex structured graph for effective motion planning within the actuator bounds. Note that the larger the size of the node, the larger is the volume of the corresponding convex set.}
    \label{fig1}
    \vspace{-0.8cm}
\end{figure}

As robots move to more complex tasks, sampling becomes non-trivial and, at times, intractable~\cite{stilman2010global,berenson2011constrained}. Extremely simple but ubiquitous tasks like rotating a crank or moving a cup of water still involve sampling from a constrained differentiable manifold~\cite{dantam2016incremental}. More specifically, continuous constraint functions need to be constructed within the configuration space $\mathcal{Q}$. Sampling-based planners operating in the ambient C-space, while task or system constraints are often defined as a set of equality conditions that specify a zero-measure subset~\cite{bonilla2015sample}. As a result, uninformed random samples have zero probability of satisfying these constraints~\cite{rodriguez2008resampl}. To the best of the authors' knowledge, no constrained sampling-based planner has utilized a planning methodology that employs a coverage estimate in its planning process. 

In this work, therefore, we take a different approach. Rather than sampling naively in all dimensions, we attempt to subsample the configuration space where feasible motions are likely to exist using a greedy compression technique~\cite{brudermuller2021trajectory}. Our central hypothesis is that effective compression enables more targeted and efficient exploration of the space. This technique also facilitates the sampling of both good (feasible) and bad (infeasible) configurations, allowing us to characterize the configuration space more thoroughly as shown in Fig.~\ref{fig1}. Once characterized, cost functions for particular planning tasks can be reformulated as a probability density function. Ultimately, convex regions are grown in the configuration space. In other words, we try to identify large, safe convex regions within the free space, which can be used to construct the GCS and plan through it efficiently. Safety, in this context, refers to maintaining configurations away from singularities while ensuring that all joint motions remain within their admissible limits.

Our main contributions are the following:
\begin{itemize}
    \item We establish a principled way to explore the configuration space of complex systems using the tensor train decomposition.
    \item We enhance planning performance within a popular convex relaxation algorithm in the context of finding the shortest paths.
    \item We verify experimentally that the identified shortest path stays within the desired configuration space and has higher manipulability than traditional sampling-based planners. 
\end{itemize}

\section{Mathematical Preliminaries}
The core idea here is to approximate a probability density function using the Tensor Train Decomposition (TTD), the goal being instantaneous retrieval of solutions. Next, we describe the concept of discovering convex regions in the C-space.

\subsection{Tensor Train Decomposition}
The TTD introduced by~\cite{oseledets2011tensor} decomposes a higher-order Tensor into low-rank tensor cores. We briefly introduce the basic notation and operations by slightly using MATLAB notation for convenience.\\

\begin{figure}[!t]
    \centering
    \includegraphics[width=1\linewidth]{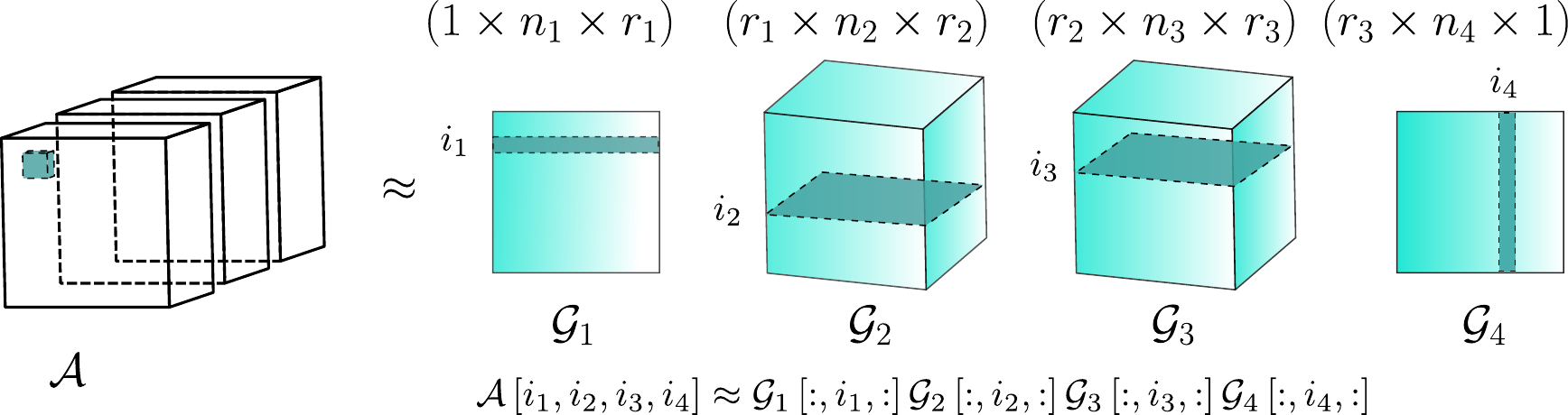}
    \caption{Representation of high-dimensional space using TT approximation. Note the decomposition into the different tensor cores.}
    \label{fig:tt_example}
    \vspace{-0.8cm}
\end{figure}

A tensor $\mathcal{A} \in \mathbb{R}^{n_1 \times n_2 \times \cdots \times n_d}$ is decomposed in tensor cores $\mathcal{G}_k \in \mathbb{R}^{r_{k-1} \times n_k \times r_k}$:
\begin{equation}
    \mathcal{A}\left[i_1, i_2, \cdots, i_d\right] \approx \mathcal{G}_1\left[:, i_1, :\right] \mathcal{G}_2\left[:, i_2, :\right] \cdots G_d\left[:, i_d, :\right]
\end{equation}
where $i_1, i_2, \cdots i_d$ denote index entries of tensor $\mathcal{A}$. $\mathcal{G}_k\left[:, i_k, :\right]$ denotes the $i_k$-th slice of tensor core $\mathcal{G}_k$. Note that $r_0 = r_d = 1$. Assuming that each tensor core has rank $r_k = r$ and dimension $n_k = n$, the required number of parameters may be estimated as $P = 2nr + \left(d-2\right)nr^2$. Thus, the number of parameters scale linearly with the dimensionality and quadratically with the rank of the cores.\\
\newline
The TTD also provides basic operations in the TT format, such as elementwise addition. Suppose two tensors $\mathcal{A}, \mathcal{B} \in \mathbb{R}^{n_1 \times n_2 \cdots \times n_d}$ have decompositions \(\mathcal{A} \approx \mathcal{G}^{\mathcal{A}}_1 \cdots \mathcal{G}^{\mathcal{A}}_d\) and \(\mathcal{B} \approx \mathcal{G}^{\mathcal{B}}_1 \cdots \mathcal{G}^{\mathcal{B}}_d\).
Then the tensor cores $\mathcal{C} = \mathcal{A} \oplus \mathcal{B}$ can be expressed with tensor cores of $\mathcal{A}$ and $\mathcal{B}$ s.t.

\begin{align}
\begin{split}
    \mathcal{G}^{\mathcal{C}}_1[:, i_1, :] &= \begin{pmatrix}
        \mathcal{G}^{\mathcal{A}}_1[:, i_1, :] & \mathcal{G}^{\mathcal{B}}_1[:, i_1, :]
    \end{pmatrix},\\
    \mathcal{G}^{\mathcal{C}}_d[:, i_d, :] &= \begin{pmatrix}
        \mathcal{G}^{\mathcal{A}}_d[:, i_d, :] \\\mathcal{G}^{\mathcal{B}}_d[:, i_d, :]
    \end{pmatrix},\\
    \mathcal{G}^{\mathcal{C}}_k\left[:, i_k, :\right] &= \begin{pmatrix}
        \mathcal{G}^{\mathcal{A}}_k\left[:, i_k, :\right] & 0 \\
        0 & \mathcal{G}^{\mathcal{B}}_k\left[:, i_k, :\right]
    \end{pmatrix}
\end{split}
\label{eq:tt_add}
\end{align}

This operation leads to an increase in rank, which may be alleviated by also utilizing the \textit{rounding operation} described in ~\cite{oseledets2011tensor}.
\subsection{PDF approximation by TT-Cross}
The TT-Cross algorithm ~\cite{ALSTTCross, OSELEDETS201070} is used to approximate high-dimensional tensors, which otherwise would become intractable to store in computer memory due to the curse of dimensionality. TT-Cross employs a greedy sampling strategy over the high-dimensional object and iteratively refines the tensor cores, with the corresponding rank updates being determined adaptively by the algorithm itself. It is therefore suitable for approximating (high-dimensional) black-box functions. In our case, TT-Cross is utilized to explore the configuration space of a robotic manipulator. For identifying certain configurations of a robot manipulator, we replace a task-specific cost function or metric with a probability density function.

\begin{figure*}[t]
    \centering
    \includegraphics[width=\textwidth]{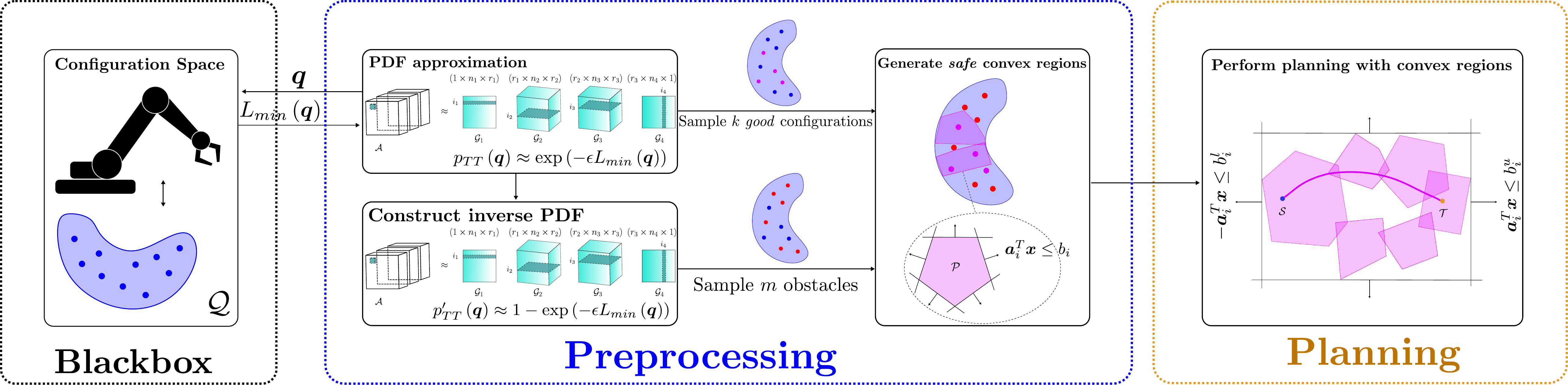}
    \caption{An overview of our TANGO algorithm is illustrated as follows. We begin by sampling the configuration space and constructing an inverse probability density function (PDF) using a chosen task metric $L_{min}$.  From this distribution, samples are drawn and classified into feasible and infeasible categories. These classifications, along with IRIS, enable the discovery of safe convex regions within the configuration space. Finally, a shortest-path search over the discovered convex sets yields the resulting trajectory from start to goal configuration.
}
\label{fig:tango_pipeline}
\end{figure*}

\subsection{Discovering Convex Sets for Planning}
We now want to describe the free C-space such that a convex characterization is possible. Iterative Regional Inﬂation by Semi-deﬁnite programming (IRIS), first introduced in~\cite{deits2014footstep}, begins by selecting a seed point in $C_{free}$ and iteratively inflates a convex polytope around it that remains entirely within the free space. Directly maximizing the volume of a convex polytope (represented as an intersection of halfspaces) is computationally intractable, as the volume of such polytopes is P-hard to compute. Instead, IRIS maximizes the volume of the largest ellipsoid that can be inscribed within the polytope as a proxy for region size.

The optimization problem is bi-convex in the decision variables defining the separating hyperplanes (halfspaces) and those defining the ellipsoid. Although not jointly convex, the problem can be efficiently solved via alternating optimization: one alternates between (i) pushing the separating hyperplanes outward to exclude nearby obstacles (the SeparatingHyperplanes step), and (ii) expanding the ellipsoid to fit maximally within the current polytope (the InscribedEllipsoid step)~\cite{deits2014footstep}. Through this iterative process, IRIS converges to a locally maximal convex region that is certified to be free of collisions.

Finding the shortest path in C-space is often approached by constructing and searching a graph, typically through sampling. To this end, local segments can be joined by shortest path queries once a dense global graph has been built. As mentioned in the introduction, another approach to constructing paths is using local optimization. This naturally raises the question of how best to balance dense sampling with effective optimization in order to approach true optimality. One way to look at this problem is through the lens of convexity. Convexity provides a unifying principle for relating both sets and functions. Once the problem is formalized as a convex program, specialized optimization tools can be employed for global optimality. One such framework that has been introduced in the recent past is the Graph of Convex Sets~\cite{marcucci2023motion,marcucci2024shortest} (GCS). We make use of a variant of GCS - the shortest path problem (SPP) - to find a geometric path inside convex regions connecting the start and target pose.    

\section{Problem Formulation}
The planning problem that we focus on is the following:
\textit{Given a start and a target configuration, and a sampling metric, we seek to identify and grow safe convex regions within compressed, low-rank tensor representation of the configuration space. These regions serve as the foundation for planning a smooth, shortest path that adheres to the system’s mechanical constraints.}

More formally, let $\mathcal{M}: \mathbb{R}^m \to \mathbb{R}_{\geq 0}$ denote a task-specific 
\emph{sampling metric}, which assigns a non-negative cost $C(q)$ to each configuration 
$\myvec{q} \in \mathbb{R}^m$. Given a start configuration $\myvec{q}_s \in \mathbb{R}^m$, a target configuration $\myvec{q}_t \in \mathbb{R}^m$, and the sampling metric $\mathcal{M}$, our 
goal is to identify a compressed representation of the feasible configuration space 
that preserves safety and tractability.  
To this end, we approximate the density of $C_{\mathrm{fes}}$
\begin{equation}
    p(q; \gamma) = \exp(-\gamma C(q)),
\end{equation}
using \emph{TT-Cross} to efficiently encode 
high-dimensional feasibility information. From this representation, we seek to 
\emph{discover convex regions} in the configuration space and embed them into a 
\emph{Graph of Convex Sets (GCS)}. 

The planning objective is then to compute a smooth, collision-free path $\pi$
\begin{equation}
    \pi: [0,1] \to \mathbb{R}^m, \quad \pi(0)=\myvec{q}_s, \;\; \pi(1)=\myvec{q}_t,
\end{equation}
that respects both the \emph{mechanical bounds} of the system and the 
\emph{geometric constraints} of the identified convex regions, 
while minimizing a cost functional (e.g., path length, energy, or task-specific metrics).

\section{Task-Specific Sampling Metrics}
The metric $\mathcal{M}: \mathbb{R}^m \to \mathbb{R}_{\geq 0}$ has to be chosen depending on the task at hand. In general, we want to minimize some cost function $\psi(\myvec{q},\myvec{\dot{q}}) \in \mathbb{R}$ that encodes a suitable performance criterion. For exposition purposes, we briefly describe two metrics that we study in the work: (i) the Yoshikawa Manipulability Metric, and (ii) a Riemannian Metric. 

The Yoshikawa manipulability metric provides a quantitative measure of how dexterous a manipulator is at a given configuration~\cite{yoshikawa1985manipulability}. It is 
defined as the volume of the velocity ellipsoid induced by the Jacobian, i.e., 
\begin{equation}
\label{eq:Yoshikawa}
C(\myvec{q}) = \sqrt{\det(\myvec{J}(\myvec{q})\myvec{J}(\myvec{q})^\top)},
\end{equation}
where $\myvec{J}(\myvec{q})$ is the manipulator Jacobian. High values of $C(\myvec{q})$ 
correspond to configurations with greater ability to generate 
end-effector motions in arbitrary directions, while low values 
indicate kinematic singularities or restricted mobility. In our 
framework, we treat $C(\myvec{q})$ as a task-specific sampling metric to 
bias the TT-Cross decomposition toward more dexterous regions of the 
configuration space. This allows us to capture manipulability-aware 
feasibility information in the compressed representation.

For the next metric, we adopt a geometry-aware singularity index \(\xi\), defined as the squared Riemannian distance between the manipulability ellipsoid \(\myvec{M}(\myvec{q}) = \myvec{J}\myvec{J}^T\) and a reference ellipsoid \(\Sigma\), i.e.,
\begin{equation}
\label{eq:Riemannian}
    \xi(\myvec{q}) = \left\| \log \left( \Sigma^{-1/2} \myvec{M} \Sigma^{-1/2} \right) \right\|_F^2\
\end{equation}
We assume that the ellipsoid \(\Sigma\) can be regarded as a hypersphere, representing the optimal configuration of a manipulator. Unlike the traditional Yoshikawa metric, which only considers the volume of the manipulability ellipsoid, \(\xi\)$(\myvec{q})$ encapsulates its full geometry—including size, shape, and orientation. This comprehensive characterization enables more accurate detection of singularities and better-informed sampling of the configuration space. Since the Yoshikawa index can remain unchanged even when the ellipsoid undergoes significant deformation, it may fail to reflect critical changes in manipulability. In contrast, \(\xi\)$(\myvec{q})$ is sensitive to such changes due to its use of a Riemannian metric that respects the affine-invariant structure of symmetric positive definite matrices~\cite{maric2021riemannian}. This makes \(\xi\)$(\myvec{q})$ a more reliable and robust tool for guiding motion planning and singularity avoidance in complex robotic systems.

\section{Tensor and Graph Optimization (TANGO)}\label{sec:method}    
Our approach can be split into two stages as elucidated in Fig.~\ref{fig:tango_pipeline}, which are required to perform path planning within safe regions of the robot configuration space.

\subsection{Tensor Train Preprocessing}
\label{subsub:preprocessing}
The preprocessing stage requires a PDF $p\left(\myvec{q};\gamma\right)$ from the \textit{blackbox} system, which in our case is a robotic manipulator. As in Shetty et al. ~\cite{shetty2024tensor}, a metric or cost function is employed to construct a PDF. Good configurations of the robot manipulator are assumed to achieve a high likelihood; singular configurations should have a small likelihood. Using TT-Cross, we approximate a high-dimensional PDF $p_{TT}$, which represents preferable configurations. By using Equation \ref{eq:tt_add}, the inverse $p'_{TT} = 1 - p_{TT}$ can be constructed in the TT format. After both PDFs are available in the TT format, we obtain \textit{initial}- and \textit{obstacle} configurations, which can be efficiently sampled in the TT format ~\cite{dolgov2020approximation}. 

For computing safe convex sets with IRIS~\cite{deits2015computing}, the obstacle configurations need to be converted to convex sets. We employ RNNDBSCAN clustering to merge obstacles into convex sets~\cite{bryant2017rnn}. Algorithm \ref{alg:preprocessing} summarizes the overall procedure.



\begin{algorithm}[t]
	\caption{Tensor Train Preprocessing}
    \label{alg:preprocessing}
    \begin{small}
		\begin{algorithmic}[1]
		    \Procedure{Preprocess}{$p$}
                \State $\mathcal{A} \gets \text{Approximate}\:p\left(\joints,\gamma\right)$ \text{using TT-Cross} once;
                \State $\Tilde{\mathcal{A}} \gets \text{Construct inverse TT-PDF using Equation \ref{eq:tt_add}}$;
                \State $\mymatrix{C}_{k} \gets \text{Draw}\:k_c\:\text{samples from}\: \mathcal{A}\:\text{using TT-Sample}$;
                \State $\mymatrix{C}_{n} \gets \text{Draw}\:n_o\:\text{samples from}\: \Tilde{\mathcal{A}}\:\text{using TT-Sample}$;
                \State $\mymatrix{C}_{k'} \gets \text{Select best}\:k'_c\:\text{configurations from}\:\mymatrix{C}_{k}$;
                \State $\mymatrix{C}_{n'} \gets \text{Select best}\:n'_o\:\text{obstacles from}\:\mymatrix{C}_{n}$;
                \State $\mathrm{OBST} \gets \text{Construct convex obstacles from}\:\mymatrix{C}_{n'}$;
                \State $\mathrm{SAFE} \gets \text{Perform IRIS using}\:\mathrm{OBST}\:\text{and}\:\mymatrix{C}_{k'}$;
            \EndProcedure
            \State return $\mathrm{SAFE}$;
	    \end{algorithmic}
 \end{small}
\end{algorithm}

\subsection{Shortest Path Planning}
The safe convex set candidates from the preprocessing stage are further pruned, since safe convex sets may reside within another convex set. These special cases won't contribute to the planning stage and introduce additional complexity during the later optimization procedure within GCS. The remaining sets are now used to construct the general graph structure for GCS. For each pair of the remaining convex sets, we check for intersections. Each intersection then represents an edge within the GCS structure.


Finally, the resulting GCS may now be used for planning (c.f. Algorithm \ref{alg:TANGO}). 

\subsection{Shortest Path Problem in Graph of Convex Sets (GCS)}

We consider the shortest path problem (SPP) formulated over a \emph{Graph of Convex Sets} (GCS). Let \( G := (\mathcal{V}, \mathcal{E}) \) be a directed graph, where:
\begin{itemize}
    \item Each vertex \( v \in \mathcal{V} \) is associated with a non-empty compact convex set \( \mathcal{X}_v \subset \mathbb{R}^n \),
    \item Each vertex also contains a continuous decision variable \( \mathbf{x}_v \in \mathcal{X}_v \),
    \item Each edge \( e = (u, v) \in \mathcal{E} \) is associated with a convex cost function \( \ell_e(\mathbf{x}_u, \mathbf{x}_v) : \mathbb{R}^n \times \mathbb{R}^n \to \mathbb{R}_{\geq 0} \cup \{\infty\} \).
\end{itemize}

A path \( p = (v_0, \dots, v_K) \) is a sequence of distinct vertices such that:
\[
v_0 = s, \quad v_K = t, \quad (v_k, v_{k+1}) \in \mathcal{E}, \quad \forall k = 0, \dots, K-1.
\]
We denote the set of edges traversed by this path as:
\[
\mathcal{E}_p := \{ (v_0, v_1), \dots, (v_{K-1}, v_K) \},
\]
and let \( \mathcal{P} \) denote the set of all valid \( s \)-\( t \) paths in the graph \( G \).

The shortest path problem over GCS is then defined as:

\begin{align}
\text{minimize} \quad & \sum_{e = (u, v) \in \mathcal{E}_p} \ell_e(\mathbf{x}_u, \mathbf{x}_v) \notag \\
\text{subject to} \quad & p \in \mathcal{P}, \notag \\
& \mathbf{x}_v \in \mathcal{X}_v, \quad \forall v \in p. 
\end{align}

Here, the decision variables are both:
\begin{itemize}
    \item the discrete path \( p \in \mathcal{P} \), and
    \item the continuous vertex configurations \( \mathbf{x}_v \in \mathcal{X}_v \).
\end{itemize}
Constraining the path to traverse vertices $\mathcal{X}_v$ inherently enables avoiding self-collisions, which may be encountered in path planning with robotic manipulators.
\subsubsection*{Edge Cost Functions}

Two common choices for edge cost functions are:

\paragraph{Euclidean Distance (Path Length)}
\begin{equation}
\ell_e(\mathbf{x}_u, \mathbf{x}_v) := \| \mathbf{x}_v - \mathbf{x}_u \|_2. \tag{2.2}
\end{equation}

\paragraph{Squared Euclidean Distance (Energy-like)}
\begin{equation}
\ell_e(\mathbf{x}_u, \mathbf{x}_v) := \| \mathbf{x}_v - \mathbf{x}_u \|_2^2. \tag{2.3}
\end{equation}

\subsubsection*{Feasibility via Convex Edge Constraints}

To enforce edge feasibility, one can define a convex constraint set \( \mathcal{X}_e \subset \mathbb{R}^{2n} \) and set:
\[
\ell_e(\mathbf{x}_u, \mathbf{x}_v) = \infty \quad \text{if } (\mathbf{x}_u, \mathbf{x}_v) \notin \mathcal{X}_e.
\]

This formulation allows us to encode motion constraints or dynamic feasibility into the edge cost.

\begin{algorithm}[t]
	\caption{Planning using TANGO}
    \label{alg:TANGO}
    \begin{small}
		\begin{algorithmic}[1]
		    \Procedure{TANGO}{$p,\myvec{q}_{start}, \myvec{q}_{goal}$}
                \State $C_{safe} \gets \text{Execute} \: 
                \mathrm{Preprocess}(p)\:\text{and prune}$ \Comment{static, initialized only once};
                \State $\mymatrix{S}\gets \text{Compute all possible intersections from}\: C_{safe}$;\Comment{static, initialized only once};
                \State $\mathrm{WP}\gets \emptyset$
            \If{$\myvec{q}_{start},\myvec{q}_{goal} \:\text{in}\: C_{safe}$}
            \State $\mathrm{GCS} \gets \text{Construct GCS using}\:\mymatrix{S}$;
            \EndIf
            \EndProcedure
            \State return $\mathrm{WP}$;
	    \end{algorithmic}
 \end{small}
\end{algorithm}

\section{Implementation Details}\label{sec:impl}
To accelerate computations, we utilize multiprocessing, where possible. We carefully select operations that are eligible for parallel computations.
\\
\newline
\textbf{IRIS}: Computing a large number of convex sets using IRIS easily becomes a computational bottleneck, especially when several convex candidates need to be computed. We assume that the initial configurations sampled by TT can be in a close neighborhood. After computing a batch of initial safe convex sets, we check if some of the remaining configurations are covered by the convex sets/polyhedra. Candidates that are already covered are then not considered for further computations. Therefore, the resulting computational complexity may be, in the best case, linear, and in the worst case, still quadratic.
\\
\newline
\textbf{Pruning}: Pruning the candidate convex sets can also, in general, result in quadratic runtimes, since checking for each possible intersection requires in total $n\left(n-1\right)/2$ combinations. To accelerate the pruning procedure, we first sort the convex set candidates according to their volume in descending order. Sets, which are contained in the sets of a bigger volume, or which are mostly covered by a set of greater volume, are removed from the potential candidate sets.

\section{Experimental Validation}
We demonstrate the effectiveness of our TANGO algorithm using $3$ distinct perspectives. Our main objective in these simulation studies is to show that our proposed framework is scalable and can be used to generate high quality trajectories for physical systems. In addition, we also elucidate the TANGO memory footprint.\\
\newline
During the illustrative example and the experiment, we utilize tunable parameters as represented in Table ~\ref{tab:tango}.

\begin{table}[t]
\label{tab:tango}
\caption{Tunable parameters in TANGO.}%
\centering%
\renewcommand{\arraystretch}{1.0}%
\begin{tabular}{>{
\raggedright}p{0.3\columnwidth}>%
{\raggedright}p{0.49\columnwidth}}
\hline
\small
Notation & Corresponding Description 
\tabularnewline
\hline 
$k_c $ &  Number of initial configurations
\tabularnewline

$k'_c $ &  Number of $k'_c$ best configurations
\tabularnewline

$n_o$ & Number of obstacle configurations
\tabularnewline
$n'_o$ & Number of $n'_o$ best obstacle configurations
\tabularnewline

\hline
$\gamma$ &  Scaling factor for cost function/metric
\tabularnewline
$n_{nswp}$ & Number of TT-Cross iterations
\tabularnewline
$n_{IRIS}$ & Number of IRIS iterations
\tabularnewline
\hline 
\end{tabular}\label{TABLE:Notations}
\end{table}
The parameters used are described in the corresponding (sub)-sections.

\subsection{Illustrative Analysis: 3-DoF Planar Manipulator}
\label{subsec:3DoF}

\begin{figure}[!t]
    \centering
    \includegraphics[width=0.7\linewidth]{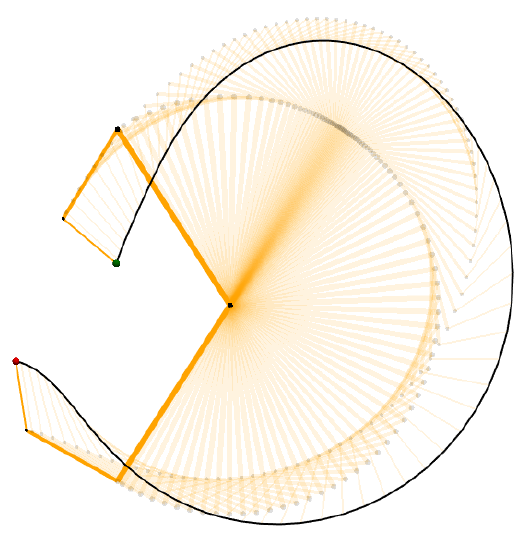}
    \caption{Executed robot trajectory from start configuration (red sphere) to goal configuration (green sphere) using TANGO. It should be noted that, due to the presence of joint limits, the geodesically shortest path in the configuration space does not necessarily correspond to the shortest path in the task (operational) space.}
    \label{fig:planar3dof_tt_trajectory}
\end{figure}

\begin{figure}[!t]
    \centering
    \includegraphics[width=0.7\linewidth]{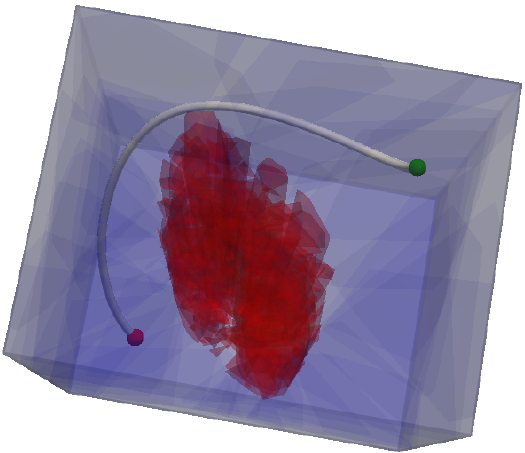}
    \caption{Safe convex sets and path planning. The blue convex sets are used for planning with TANGO, and the red convex sets represent obstacles in the configuration space. The final path is shown in white.}
    \label{fig:planar3dof_convex_sets}
\end{figure}

\begin{figure}[!t]
    \centering
    \includegraphics[width=1.0\linewidth]{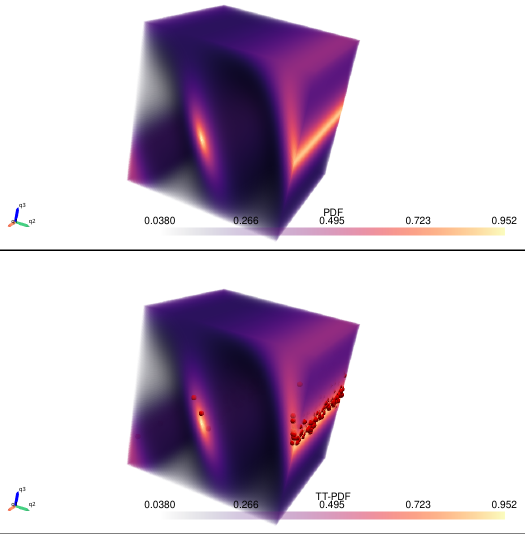}
    \caption{Original pdf (top) and TT-Cross approximation (bottom) for the Yoshikawa metric. The red points (bottom picture) denote samples taken from the TTD. Regions with a higher probability are sampled more likely. For illustrative and visualization purposes, we fixed the scaling parameter value to $\gamma = 0.1$.}
    \label{fig:planar3dof_tt_yoshikawa}
    \vspace{-0.8cm}
\end{figure}

To evaluate the effectiveness of the proposed framework, we begin with an illustrative case study involving a simple $3$ DoF planar robotic manipulator (taken from Lück et al. ~\cite{luck1993self}), designed to move its end-effector in a $2$D workspace. First, we need to decide on which metric is to be utilized for constructing a PDF used for sampling. We consider the classic Yoshikawa metric and the Riemannian metric (c.f Equation ~\ref{eq:Yoshikawa},~\ref{eq:Riemannian}).
We show the PDFs and their corresponding approximations in Figure ~\ref{fig:planar3dof_tt_yoshikawa} and Figure ~\ref{fig:planar3dof_tt_riemannian}. The planar 3DoF manipulator is a non-escapable singular configuration (c.f. ~\cite{luck1993self}) when fully stretched. The singular configurations appear as low PDF values occuring in roughly in the center of the Figures. After sampling from both PDFs, we clearly see that it is likely to happen to sample near-singular configurations when constructing a PDF using Equation ~\ref{eq:Yoshikawa}. Therefore, from here on, we will just consider the Riemmanian metric.
\\
\newline
Continuing our illustrative example we demonstrate TANGO's core components—tensor-based feasibility modeling, convex region generation, and graph-based trajectory planning—under controlled conditions.

We discretize each joint of the $3$ DoF manipulator into 128 bins, resulting in a dense 3D tensor \( \mathcal{A} \in \mathbb{R}^{128 \times 128 \times 128} \), which encodes a task-specific feasibility metric. To enhance computational tractability, we reshape this tensor into a 7D structure \( \mathcal{A}' \in \mathbb{R}^{8 \times 8 \times \cdots \times 8} \), enabling efficient compression via TT-Cross decomposition.

\begin{figure}[!t]
    \centering
    \includegraphics[width=1.0\linewidth]{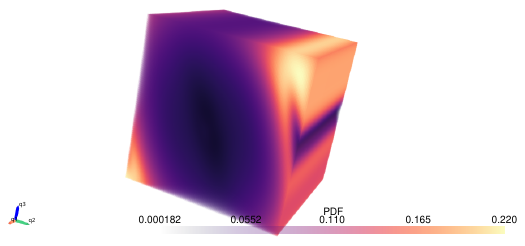}
    \caption{Original PDF constructed using the Riemannian metric ($\gamma = 1.0$). The Riemannian metric results in comparatively more favorable manipulability configurations, whereas the Yoshikawa metric induces a more restricted spatial domain characterized by high-manipulability regions.}
    \label{fig:orig_riemannian}
\end{figure}

\begin{figure}[!t]
    \centering
    \includegraphics[width=1.0\linewidth]{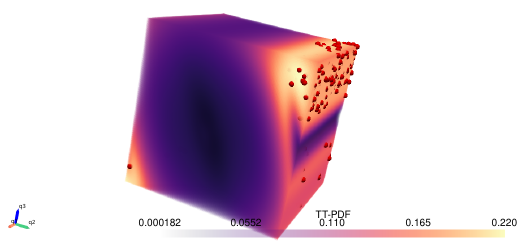}
    \caption{TT-Cross approximation of the PDF. The red points here denote samples taken from the TTD. Regions with a higher probability are sampled more likely.}
    \label{fig:planar3dof_tt_riemannian}
    \vspace{-0.8cm}
\end{figure}

Figures~\ref{fig:orig_riemannian} and~\ref{fig:planar3dof_tt_riemannian} show both the full feasibility tensor and its TT-approximated counterpart. To generate the inverse feasibility field, we use the element-wise TT subtraction operation:
\[
\mathcal{A}_{\text{inv}} = \mathds{1} \ominus \mathcal{A}',
\]
where \( \mathds{1} \) is a unit-valued tensor and \( \ominus \) denotes element-wise subtraction in TT format.

For environment modeling, we sample \(1.5 \times 10^4\) configurations, including both feasible and obstacle-occupied points. From these, we select the top \(10^4\) obstacle configurations and 500 high-quality feasible samples based on the task-specific metric. The obstacle configurations are clustered using RNN-DBSCAN, producing convex obstacle regions, while safe regions are extracted using the IRIS algorithm and pruned as described in Section~\ref{sec:impl}.

The final Graph of Convex Sets (GCS) is constructed by identifying intersecting convex regions, enabling shortest-path planning through safe corridors. Figure~\ref{fig:planar3dof_convex_sets} illustrates a planned trajectory across convex sets. This example assumes no self-collision, focusing instead on validating TANGO’s ability to identify and traverse meaningful configuration space regions under task-specific metrics.

\subsection{Scalability and Trajectory Generation for Panda Arm}
\label{subsec:panda4dof}

\begin{figure*}[h]
    \centering
    \includegraphics[width=\textwidth]{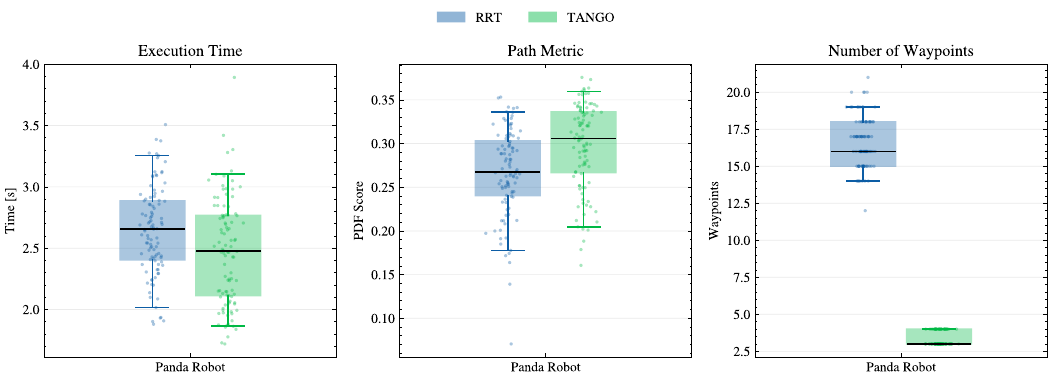}
    \caption{Performance evaluation of TANGO and RRT. Execution times of executed trajectories (left), PDF-Score (middle), and number of resulting waypoints (right). The execution times of the generated trajectories are, on average higher for TANGO, while having a better PDF Score (higher values indicate better performance). TANGO requires in general, fewer waypoints, leading to smooth trajectories after further refinement.}
    \vspace{-0.6cm}
\end{figure*}

To demonstrate the scalability and trajectory quality of our proposed method, we now evaluate its performance on a high-dimensional manipulator model—a 7-DoF Franka Emika Panda robot~\cite{haddadin2024franka}. To reduce experimental complexity while still retaining meaningful motion planning challenges, we lock joints $5, 6,$ and $7$, resulting in a $4$-DoF configuration space. Extending the framework to seven degrees of freedom (7DoF) is expected to increase computational complexity, as the safe convex sets—represented as polyhedra—will, in general, exhibit greater geometric complexity. Consequently, in the present work, we focus on a lower-dimensional, simplified problem setting, with the intention of subsequently generalizing the approach to systems with a larger number of degrees of freedom in future research.

We discretize the $4$D space into \(128\) equally spaced bins per joint, forming a tensor $\mathcal{A} \in \mathbb{R}^{128 \times 128 \times 128 \times 128}$. For efficient high-dimensional compression, the tensor is reshaped into a $14$D tensor $\mathcal{A}' \in \mathbb{R}^{4 \times 4 \times \cdots \times 4}$, which is then approximated using the TT-Cross decomposition technique. It should be acknowledged that alternative choices for the tensor dimensionality are feasible, and that the determination of both an optimal number of dimensions \cite{novikov2015tensorizing} and an appropriate ordering of these dimensions \cite{tichavsky2025optimizing} lies beyond the scope of the present study. The specific dimensional configuration employed in this work was determined empirically. TT-Cross approximates the PDF within $n_{nswp} = 30$ iterations. We construct the PDF by scaling the Riemannian Metric \eqref{eq:Riemannian} with a factor $\gamma = 0.1$.

Following the feasibility approximation, we construct the inverse feasibility tensor and sample $n_o = k_c = 5 \times 10^4$ configuration points. From these, we select the best $k'_c = 10^3,\: n'_o = 10^4$ samples from each set based on the Riemannian metric. Configurations with self-collisions among the initial candidates are reclassified as obstacles. Convex obstacle regions and safe convex sets are then extracted using the same pipeline as in Section~\ref{subsec:3DoF}. Figure~\ref{fig1} shows the convex regions connected using a graph. We perform $n_{IRIS} = 2$ IRIS iterations for the convex set computations. This ensures that the convex sets stay in the neighborhood of their initial configurations.

To evaluate trajectory quality, we compare TANGO to the standard RRT planner. We randomly select $100$ start–goal pairs with large joint-space distances to ensure non-trivial planning tasks. Both planners generate collision-free paths, which are subsequently refined using TOPPRA~\cite{pham2014general} to yield smooth, dynamically feasible trajectories. We evaluate each trajectory and measure the worst-case possible PDF-Score.
This gives us an empirical lower bound estimate of the performance of each of the approaches.

\begin{figure*}[t]
    \centering
    \includegraphics[width=1\linewidth]{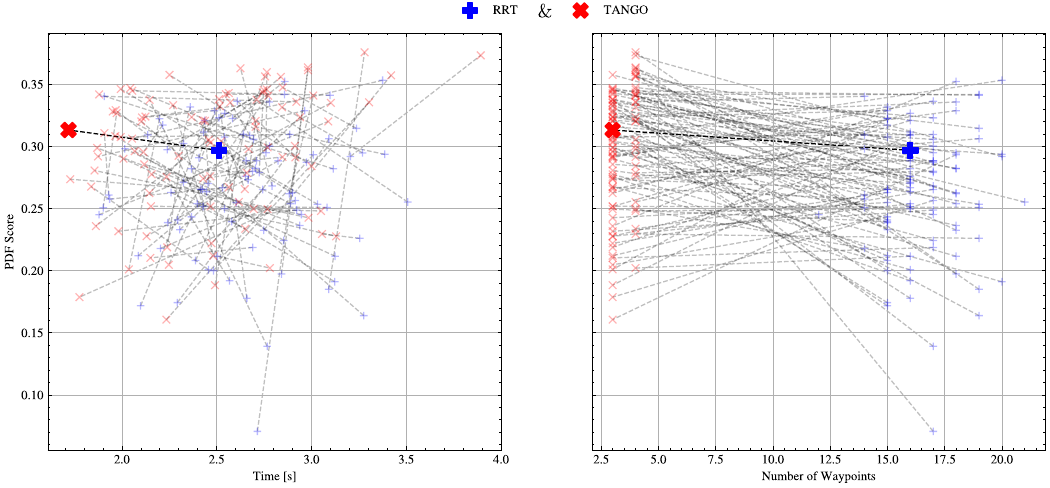}
    \caption{Comparison between a naive RRT and TANGO. The plots show that our Riemannian sampling strategy in TANGO leads to approximate convergence in less time (left) and fewer waypoints (right) for different runs.}
    \label{fig:tango}
    \vspace{-0.6cm}
\end{figure*}

As shown in Figure~\ref{fig:tango}, TANGO consistently produces more structured and concise plans—typically requiring only 3–4 waypoints before refinement—resulting in significantly smoother trajectories than those produced by RRT. In addition, our approach yields a higher minimum PDF-score during each trajectory on average. This experiment highlights TANGO’s ability to scale to higher-dimensional systems while maintaining both computational efficiency and trajectory quality.

\subsection{Time \& Memory Footprints}

To further evaluate the practicality of TANGO, we 
analyze its computational efficiency with respect to two 
critical bottlenecks in robotics software: execution time and memory consumption. These metrics are particularly 
important in resource-constrained systems, where planners must operate under strict latency requirements while maintaining a small memory footprint. By systematically profiling both dimensions, we assess how well our method scales in comparison to existing baselines and whether it 
remains viable for real-world deployment on embedded or 
low-power platforms.

Table~\ref{tab:ttcross-comparison} compares the memory footprint and execution time of the TT-Cross algorithm when applied to tensors with original versus reshaped dimensions, averaged over 15 trials. The reshaped tensor ($4^{14}$ dimensions) demonstrates a significantly reduced memory footprint compared to the original tensor ($128^4$ dimensions). Specifically, the mean number of parameters drops from approximately $4.36 \times 10^4$ to $7.11 \times 10^3$, with consistently lower variability, as indicated by a smaller standard deviation. Execution time also improves substantially: the mean time decreases from $27.79$ seconds for the original tensor to just $5.53$ seconds for the reshaped version. Minimum and maximum values in both metrics further confirm this trend, highlighting that the reshaped tensor yields faster and more memory-efficient performance across all trials.

\begin{table}[h!]
\centering
\caption{Comparison of TT-Cross memory footprint and execution time for original and reshaped tensor dimensions over $15$ trials.}
\label{tab:tt_comparison}
\begin{tabular}{l|r|r}
\toprule
\textbf{Metric} & \textbf{Original Dim.} & \textbf{Reshaped Dim.} \\
& $128^4$ & $4^{14}$ \\
\midrule
\multicolumn{3}{c}{\textbf{Memory Footprint (Parameters)}} \\
\midrule
Mean ($\mu$) & $\num{4.36e+04}$ & $\num{7.11e+03}$ \\
Std Dev ($\sigma$) & $\num{2.25e+04}$ & $\num{1.70e+03}$ \\
Min & $\num{1.08e+04}$ & $\num{5.28e+03}$ \\
Max & $\num{6.66e+04}$ & $\num{1.19e+04}$ \\
\midrule
\multicolumn{3}{c}{\textbf{Execution Time (s)}} \\
\midrule
Mean ($\mu$) & $\num{27.7924}$ & $\num{5.5336}$ \\
Std Dev ($\sigma$) & $\num{2.6981}$ & $\num{0.5166}$ \\
Min & $\num{20.7543}$ & $\num{4.6642}$ \\
Max & $\num{33.6748}$ & $\num{6.4782}$ \\
\bottomrule
\end{tabular}
\label{tab:ttcross-comparison}
\end{table}

\section{Conclusion and Future Work}
This work presents a principled and efficient approach to motion planning by leveraging a greedy compression technique to focus sampling within promising regions of the configuration space. By targeting areas where feasible motions are more likely, our method avoids the inefficiencies of uniform sampling across all dimensions and enables a more thorough characterization of both feasible and infeasible configurations. This, in turn, allows us to reformulate task-specific cost functions as probability density functions and grow large, safe convex regions within the free space.

These convex regions form the foundation for constructing a Graph of Convex Sets (GCS), enabling fast and reliable planning. Our contributions demonstrate the effectiveness of this approach in improving planning performance, particularly in convex relaxation-based shortest path problems. Through experimental validation, we also confirmed that the computed paths remain within the intended feasible regions, further supporting the practical value of our method.

Although task-space obstacles are not explicitly addressed in the present work, a direct extension is readily attainable. Specifically, one may construct an additional tensor train either from Signed Distance Functions (SDFs)~\cite{oleynikova2016signed, brudermuller2021trajectory} or from distance functions defined in the manipulator’s configuration space and modeled as a probability density function (PDF). Then the two tensor trains can be combined in the preprocessing step. Further research is necessary to extend the proposed methodology to systems of higher dimensionality. In particular, the choice of the metric employed for constructing the probability density function (PDF), as well as the overall dimensionality of the tensor representation, is expected to substantially influence both the memory requirements and the attainable approximation accuracy. Further, topological connections could be analyzed between special manifolds in the configuration space and the tensor compressions. Lastly, the trajectory smoothness can be improved by considering spline segments to join the convex sets during planning.    

\bibliographystyle{IEEEtran}
\bibliography{bibliography}

\end{document}